\documentclass[sigconf]{acmart}
\AtBeginDocument{%
  }

\setcopyright{acmlicensed}
\copyrightyear{2025}
\acmYear{2025}
\setcopyright{acmlicensed}\acmConference[ICMR '25]{Proceedings of the 2025 International Conference on Multimedia Retrieval}{June 30-July 3, 2025}{Chicago, IL, USA}
\acmBooktitle{Proceedings of the 2025 International Conference on Multimedia Retrieval (ICMR '25), June 30-July 3, 2025, Chicago, IL, USA}
\acmDOI{10.1145/3731715.3733484}
\acmISBN{979-8-4007-1877-9/2025/06}
\usepackage{multirow}

\begin{document}

\title{Mask-aware Text-to-Image Retrieval: Referring Expression Segmentation Meets Cross-modal Retrieval}


\author{Li-Cheng Shen}
\affiliation{%
  \institution{National Taiwan University}
  \city{Taipei}
  \country{Taiwan}
}
\email{r13922098@csie.ntu.edu.tw}

\author{Jih-Kang Hsieh}
\affiliation{%
  \institution{National Taiwan University}
  \city{Taipei}
  \country{Taiwan}
}
\email{r13922a03@csie.ntu.edu.tw}

\author{Wei-Hua Li}
\affiliation{%
  \institution{National Taiwan University}
  \city{Taipei}
  \country{Taiwan}
}
\email{d12922009@csie.ntu.edu.tw}

\author{Chu-Song Chen}
\affiliation{%
  \institution{National Taiwan University}
  \city{Taipei}
  \country{Taiwan}
}
\email{chusong@csie.ntu.edu.tw}


\begin{abstract}
Text-to-image retrieval (TIR) aims to find relevant images based on a textual query, but existing approaches are primarily based on whole-image captions and lack interpretability. Meanwhile, referring expression segmentation (RES) enables precise object localization based on natural language descriptions but is computationally expensive when applied across large image collections. To bridge this gap, we introduce Mask-aware TIR (MaTIR), a new task that unifies TIR and RES, requiring both efficient image search and accurate object segmentation. To address this task, we propose a two-stage framework, comprising a first stage for segmentation-aware image retrieval and a second stage for reranking and object grounding with a multimodal large language model (MLLM). 
We leverage SAM 2 to generate object masks and Alpha-CLIP to extract region-level embeddings offline at first, enabling effective and scalable online retrieval. 
Secondly, MLLM is used to refine retrieval rankings and generate bounding boxes, which are matched to segmentation masks. 
We evaluate our approach on COCO and D$^3$ datasets, demonstrating significant improvements in both retrieval accuracy and segmentation quality over previous methods. Our code is available at https://github.com/AI-Application-and-Integration-Lab/MaTIR.

\end{abstract}





\maketitle

\section{INTRODUCTION}

TIR \cite{faghri_vse_2017, li_visual_2019, radford_learning_2021, jia_scaling_2021} involves retrieving images from natural language descriptions. TIR can lack explainability because it retrieves relevant images without indicating where the concept occurs. On the other hand, RES \cite{ding_vision-language_2021, lai_lisa_2024, shen_aligning_2024, chen_sam4mllm_2025} identifies and segments specific objects within an image based on a description (referring expression), locating and outlining the precise boundaries within the image. In this work, we explore a new task, 
MaTIR, where relevant images are retrieved along with associated masks. For RES, our approach handles an image gallery for visual search, rather than just segmenting a specific image. For TIR, our method not only retrieves relevant images but also identifies the precise regions where the objects are located. This enhances explainability and allows the region masks applicable for other usage. 

The straightforward approach of applying RES to all images in a database is impractical due to the high computational cost. An alternative is leveraging TIR models like CLIP \cite{radford_learning_2021} to retrieve candidate images before performing RES. However, these models focus on aligning global image and text representations and are optimized for retrieval based on whole-image captions \cite{chen_microsoft_2015, plummer_flickr30k_2015}, leading to suboptimal performance for object-level retrieval in MaTIR. A related study \cite{levi_object-centric_2023} 
retrieves images containing objects based on category names from object detection datasets \cite{lin_microsoft_2014}, which we show to be insufficient for querying with 
complex textual expressions.


To tackle these challenges, we propose a two-stage method 
for solving the MaTIR problem. 
First, a segmentation-aware method is introduced for efficient image retrieval. 
Second, MLLMs are used for reranking and bounding box-level comprehension. We compare our method to other baseline methods including retrieval using the whole-image embedding and segmentation using existing RES. 

The main contributions of our work are as follows: First, we introduce a novel task, MaTIR, which unifies visual mask grounding with text-to-image retrieval. Second, we propose a method to handle MaTIR, which can effectively retrieve object regions based on natural language descriptions. Third, our method requires no training, which is generally applicable to open-vocabulary objects. Our work can serve as a base for future study toward this direction.

\begin{figure*}[t]
  \centering
  \includegraphics[width=0.7\linewidth]{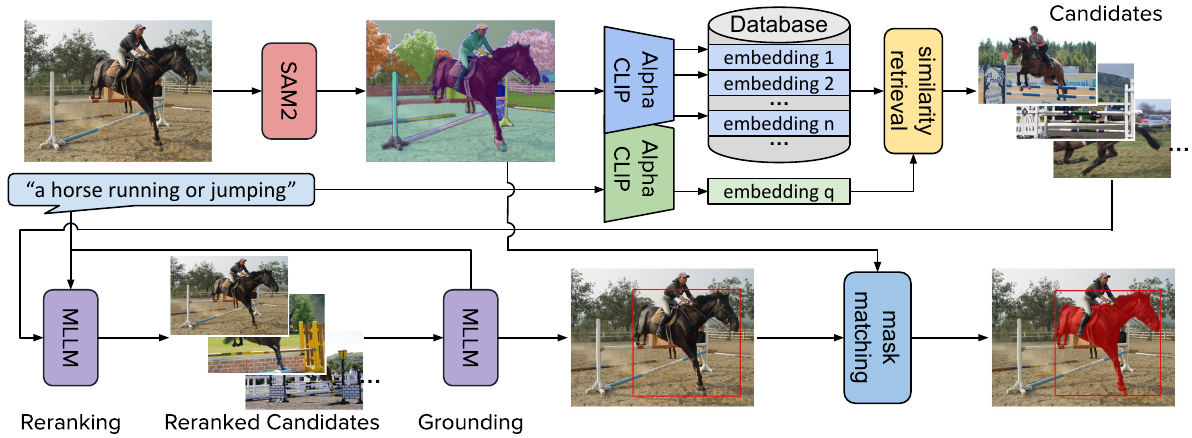}
\caption{Illustration of our method for tackling MaTIR. During offline indexing, SAM 2 \cite{ravi_sam_2024} produces mask-level region proposals, and Alpha-CLIP \cite{sun_alpha-clip_2024} extracts a visual embedding for each mask. At query time, the textual object is encoded, and a nearest neighbor search retrieves the top candidates. An MLLM then reranks the retrieved images to improve retrieval accuracy. Finally, the MLLM localizes the target objects with bounding boxes, and mask matching is used to select the output masks.}
  \Description{Illustration of our proposed framework for tackling the MaTIR task.}
  \label{fig:pipeline}
\end{figure*}

\section{RELATED WORK}
We briefly review TIR and RES in the following.

\noindent\textbf{TIR} is a long-standing task 
focusing on retrieving relevant images from an image gallery based on textual descriptions, typically evaluated on image-caption datasets \cite{chen_microsoft_2015, plummer_flickr30k_2015}. While early studies aimed at improving feature extraction and alignment, recent advancements have achieved remarkable performance through vision-language pre-training (VLP) with large-scale image-text datasets. TIR models can be categorized into single-stream and dual-stream architectures. In single-stream methods, images and texts are processed together within the model to enable cross-modal interaction. This includes earlier methods such as SCAN \cite{lee_stacked_2018} and MMCA \cite{wei_multi-modality_2020}, and recent VLP-based models such as UNITER \cite{chen_uniter_2020} and ViLT \cite{kim_vilt_2021}. However, single-stream methods require evaluating each query-image pair at query time, leading to substantial computation overhead. This makes them impractical for large-scale visual search. 

In contrast, dual-stream methods encode visual and textual features independently, and image-text similarity 
is computed between their output embeddings. This design enables offline indexing of images, allowing for efficient retrieval at query time by encoding only the text 
for a nearest-neighbor search. Examples of early dual-stream methods include DeViSE \cite{frome_devise_2013}, VSE++ \cite{faghri_vse_2017}, and VSRN \cite{li_visual_2019}. More recent VLP-based dual-encoder models include CLIP \cite{radford_learning_2021}, ALIGN \cite{jia_scaling_2021}, CoCa \cite{yu_coca_2022}, and SigLIP \cite{zhai_sigmoid_2023}. In this paper, we adopt a dual-stream framework to enable scalable text-to-image retrieval. However, models like CLIP lack the region-level granularity required for MaTIR. To address this limitation, we introduce a segmentation-aware embedding approach, which significantly enhances retrieval performance.
\\

\noindent\textbf{RES} aims to segment a target object in an image based on a natural language description. Unlike open-vocabulary segmentation \cite{xu_side_2023, chen_open-vocabulary_2024} accepting only category names, RES uses sentences as inputs for segmentation. By utilizing transformer architectures to model the interactions between visual and textual information, VLT \cite{ding_vision-language_2021} uses query generation and balance modules to adaptively select relevant queries, and LAVT combines textual features and visual features through a pixel-word attention module. X-Decoder \cite{zou_generalized_2023} develops a generalized decoding model for open-vocabulary segmentation and RES, while APE \cite{shen_aligning_2024} addresses the granularity gap between pixel-level tasks for universal grounding and segmentation. Leveraging the advanced cross-modality capabilities of MLLMs, LISA \cite{lai_lisa_2024} and LISA++ \cite{yang_lisa_2024} introduce a <SEG> token for mask decoding to enable RES, and SAM4MLLM \cite{chen_sam4mllm_2025} integrates SAM \cite{kirillov_segment_2023} into MLLMs through an active querying approach.

These methods are typically evaluated on datasets such as the RefCOCO series \cite{yu_modeling_2016, mao_generation_2016}, which assume that the referred object always exists in the given image. As this setting limits the practical applicability of RES methods, recent datasets are designed to relax this constraint. gRefCOCO \cite{liu_gres_2023} and Ref-ZOM \cite{hu_beyond_2023} consider expressions corresponding to no object or multiple objects. GRD \cite{wu_advancing_2023} segments objects across a group of images, while the object may or may not be present.
However, the images per query are few in these datasets and they still require RES models to process each image. 
Our MaTIR task aims to account for scalability, extending RES to operate efficiently over an image gallery. Our approach is compatible with existing RES models, 
while we demonstrate a simple yet effective approach by leveraging the strong capabilities of MLLMs with bounding box comprehension capabilities \cite{bai_qwen-vl_2023, bai_qwen25-vl_2025}. 

\section{METHODOLOGY}


Our method consists of two stages: \textit{segmentation-aware TIR} and \textit{MLLM-based reranking \& grounding}. Let $\mathbf{\Pi}=\{I_1,\cdots, I_N\}$ be a gallery containing $N$ images. For all $I_i$ in $\mathbf{\Pi}$, we offline extract and store its feature in the database $\mathbf{V}$ for retrieval. Then, given query text $q$ in the retrieval phase, we perform $k$-nearest neighbor search in the feature-representation space for $q$ at first, and then use MLLM-based reranking followed by grounding to complete the MaTIR task. An overview of our approach is given in Fig.~\ref{fig:pipeline}.

\subsection{Segmentation-aware TIR}
Common feature extraction solutions for TIR (e.g., CLIP \cite{radford_learning_2021}, ALIGN \cite{jia_scaling_2021}, SigLIP \cite{zhai_sigmoid_2023}) align only whole-image-based embedding with the text embedding, which excel at recognizing high-level concepts within images but may struggle with finer details requiring precise pixel-level analysis. To tackle this issue, we use SAM 2 \cite{kirillov_segment_2023, ravi_sam_2024} to generate mask-level region proposals at first, yielding multiple region masks for every image $I_i\in \mathbf{\Pi}$. Suppose $\mathcal{N}_i$ masks are obtained for image $i$, denoted as $M_{ij}$, $1\leq j \leq \mathcal{N}_i$ and $1\leq i \leq N$. 

To extract object-level instead of whole-image-level embedding, we use Alpha-CLIP \cite{sun_alpha-clip_2024}. 
Alpha-CLIP extends the CLIP model with region awareness by incorporating an auxiliary alpha channel, aligning regional representations in CLIP's feature space. It takes an object mask together with its belonging image, $(M_{ij},I_i)$, as input. Therefore, the feature embedding for a masked object depends on both the object itself and the contextual background information of the image. We input each mask along with the image into the image encoder of Alpha-CLIP, yielding a set of visual embeddings, 
$\textbf{V}=\{\textbf{v}_{ij}\in \mathbb{R}^{d}|1\leq i \leq N, 1\leq j \leq \mathcal{N}_i\}$, for the gallery images, with $d$ the embedding dimension.

At querying, the textual object query is encoded by the text encoder of Alpha-CLIP into $\textbf{q} \in \mathbb{R}^d$, and for each image $I_i$ the matching score is given by the maximum cosine similarity,
\begin{equation}
S_i = \max_{j \in [1, \mathcal{N}_i]} \frac{\textbf{q} \cdot \textbf{v}_{ij}}{\|\textbf{q}\| \|\textbf{v}_{ij}\|}.
\end{equation}
Suppose that top $N_C$ images are retrieved at this stage. 

\subsection{MLLM-based Reranking \& Grounding}
Although Alpha-CLIP can align texts with masked objects in a background image within a unified embedding space, it can only handle relatively simple text. Hence, we adopt it for a coarse search of relevant images. To further refine the retrieved results, we incorporate MLLMs for reranking, leveraging their ability to understand fine-grained semantic details for more accurate ranking. MLLMs simultaneously take both images and texts as inputs.
We prompt the MLLM with a true-or-false question to obtain a relevance score for each image, similar to previous approaches \cite{nogueira_document_2020, lin_universal_2024}.

\noindent\textbf{Reranking}: Given a textual object query \textit{<obj>} and a candidate image \textit{<img>}, we construct the following prompt for the MLLM: "\textit{<img>Does this image contain the described object? Answer True or False.}\textit{\textbackslash nObject description: <obj>}". From the model's response, we extract the output logits corresponding to the "True" and "False" tokens, denoted as $z^{True}$ and $z^{False}$, respectively. The final relevance score $S'$ is computed as
\begin{equation}
S' = \frac{\exp(z^{True})}{\exp(z^{True}) + \exp(z^{False})}.
\end{equation}
The $N_C$ images retrieved above are then re-ordered, and we keep only the top $N_K$ ($N_K<N_C$) candidates after reranking.
This step enhances the initial retrieval by leveraging multimodal reasoning, leading to a more accurate ranking of the retrieved images.

\noindent\textbf{Grounding and Mask Matching}:
Recent progress in MLLMs allows them to
access local semantic relevance 
and generate bounding box predictions for objects.
To enhance localization accuracy, we leverage MLLMs again, specifically ones with object grounding capabilities.
Given a textual object query \textit{<obj>} and an image \textit{<img>} from the top-$N_K$ 
reranked results, 
the MLLM generates the coordinates of the bounding box in the model’s response using the following prompt: "\textit{<img>Locate the described object, output the bbox coordinates in JSON format.}\textit{\textbackslash nObject description: <obj>}".
We then select the mask with the  
highest matching score 
for the given image. Specifically, each mask generated by SAM 2 has 
a bounding box, 
and we use the intersection-over-union (IoU) 
between SAM 2 and MLLM-generated bounding boxes as the matching scores. The 
selected masks then serve as the 
outputs for MaTIR.

\noindent\textbf{Evaluation Metric}:
Because MaTIR is a new task that has not been well studied before, we need to introduce a metric for its performance evaluation. 
Note that MaTIR can be viewed as retrieving segmented objects in the image gallery. 
In image-level retrieval, we adopt the commonly used mAP@50 as the performance measure. 
In the object-retrieval level, we extend mAP@50 by defining an instance as correct if the retrieved image contains the object and the predicted mask matches one ground-truth mask with an IoU of 0.5 or higher. We refer to this metric as mAP@50@50, which is used for performance evaluation of the final goal of MaTIR.

\section{EXPERIMENTS}

\textbf{Datasets}: 
For evaluating MaTIR, the datasets must meet these requirements: (1) RES Support: Mask annotations should be provided for objects described in natural language. (2) Complete Annotation: 
All images should be annotated for each object, ensuring verification of its presence or absence.


We first adopt \textbf{COCO} \cite{lin_microsoft_2014} to verify the basic capability of our approach, which has been utilized for object-level image retrieval in \cite{levi_object-centric_2023}. We follow the setup of \cite{levi_object-centric_2023} by using its validation set of 5,000 images as the image gallery, with its 80 category names serving as object queries. However, COCO offers only category names as labels rather than 
complex textual descriptions. To address 
this limitation, we use also the \textbf{D$^3$} dataset \cite{xie_described_2023}, which was originally designed for the Described Object Detection (DOD) task, in our experiments. 
D$^3$ provides both description-based mask annotations and complete annotations, fulfilling the requirements for MaTIR. We utilize the full set of 10,578 images as the image gallery and its 422 fully-annotated object descriptions as retrieval queries.

Note that our method is zero-shot and needs no training. Thus, the datasets above are only for the performance-evaluation purpose.

\noindent\textbf{Implementation Details}:
For SAM 2, we use the Hiera-L model, and set a confidence threshold of 0.5 and NMS threshold of 0.7. For Alpha-CLIP, we use their ViT-L/14 model finetuned on GRIT-20M dataset, and the embedding dimension $d=768$. When encoding textual queries, we ensemble over the 7 best CLIP prompts \cite{radford_learning_2021}. We use $N_C=100$ and $N_K=50$ in the retrieval results. For both reranking and object grounding, we use Qwen2.5-VL 7B \cite{bai_qwen25-vl_2025} as the MLLM, which is able to comprehend local object information and output bounding box coordinates with texts. 

\subsection{Results and Discussion}
\noindent\textbf{Image-level retrieval performance}: We compare with directly leveraging \textit{CLIP} \cite{radford_learning_2021}, \textit{Dense-CLIP}, and \textit{Cluster-CLIP}. Dense-CLIP and Cluster-CLIP are adopted from \cite{levi_object-centric_2023}. Dense-CLIP modifies the final multi-head attention layer of CLIP’s image encoder with ResNet backbones to extract dense patch embeddings, while Cluster-CLIP further applies an aggregation module. We evaluate both methods based on CLIP with ResNet50x64 backbone, along with K-Means clustering with 25 clusters for the aggregation module.

Table \ref{tab:retrieval} shows the image-level retrieval performance. First, Dense-CLIP and Cluster-CLIP achieve performance gains over CLIP on COCO, as observed in \cite{levi_object-centric_2023}, where they provided the explanation that global embeddings of CLIP struggle with retrieving small objects or complex images, while dense embeddings capture local semantics. However, both methods exhibit a performance drop on D$^3$, indicating that patch-based embeddings lose 
intricate contextual information, which is essential for retrieving objects described by complex language expressions (e.g., \textit{"outdoor dog led by rope"}). 

As can be seen in Table \ref{tab:retrieval}, both our coarse retrieval ("Ours w/o reranking") and fine retrieval ("Ours") outperform the compared approaches. The coarse search performs more favorably than previous approaches by 5.5 points on COCO and 7 points on D$^3$, and the reranked fine search leads to significant performance gains of 9.4 points on COCO and 20.3 points on D$^3$. The results demonstrate that incorporating SAM 2 with Alpha-CLIP embeddings captures fine-grained contextual semantics, which is crucial for MaTIR. The multimodal understanding capabilities of MLLMs can further refine retrieval results for substantially enhancing the retrieval accuracy.

\noindent\textbf{Object-level retrieval \& segmentation performance}: After retrieving the $N_K$ images, our method leverages MLLMs for bounding box generation, combined with SAM 2 mask matching. Our method is thus zero-shot and applicable to open-vocabulary situations.

To 
process the retrieved $N_K$ images, it is also natural to utilize existing RES methods. In the experiments, we compare our method with existing RES solutions, \textit{LISA++} \cite{yang_lisa_2024}, \textit{SAM4MLLM} \cite{chen_sam4mllm_2025}, and \textit{APE} \cite{shen_aligning_2024}. For LISA++, we use their 7B model for evaluation, and take the first mask in the MLLM response as the output. For SAM4MLLM, we use their model based on LLaVA1.6 8B \cite{liu_llava-next_2024} for evaluation. For APE, we use their APE-D model for evaluation, and take the mask with the highest probability as the output. For a fair comparison, each RES method is applied after our segmentation-aware TIR and MLLM reranking (\textit{i.e.}, their TIR performance is the same as the "Ours" row in Table \ref{tab:retrieval}). The comparison is provided in Table \ref{tab:res} using the mAP@50@50 metric.






As shown in Table \ref{tab:res}, LISA++ performs the worst. APE and SAM4MLLM perform 
best among RES methods on COCO and D$^3$, respectively. However, they are all inferior to 
our approach. The reason for this is that most existing RES methods are limited in their generalization capability to unseen datasets or objects because they are trained on specific RES datasets. On the other hand, MLLM's ability to locate objects by outputting their bounding box coordinates is more generalizable. Therefore, our method, which leverages MLLM combined with the general segmentation model SAM 2, performs more favorably on the MaTIR task.



\subsection{Ablation Study}

\noindent\textbf{Region-level embedding}: 
We employ Alpha-CLIP to extract region-level feature embeddings. As an alternative, we consider another method for obtaining region embeddings: cropping the bounding boxes surrounding SAM 2 masks, resizing them, and feeding them into CLIP’s image encoder. For a pure comparison, we omit the reranking step and report the retrieval performance of this approach alongside our original Alpha-CLIP method in Table \ref{tab:ablation1}. The results show a performance drop, indicating that cropping the region reduces contextual background information crucial for recognition and semantic understanding. This highlights the advantage of Alpha-CLIP’s region-aware feature extraction in our method.


\begin{table}[!t]
  \caption{Comparison of methods for the TIR stage.} 
  \renewcommand{\arraystretch}{0.45}
  \label{tab:retrieval}
  \resizebox{0.9\columnwidth}{!}{
    \begin{tabular}{lcc}
      \toprule
      Method & COCO (mAP@50)& D$^3$ (mAP@50)\\
      \midrule
      CLIP$_{\text{ResNet50x64}}$ \cite{radford_learning_2021} & 71.00 & 33.23 \\
      CLIP$_{\text{ViT-L/14}}$ \cite{radford_learning_2021} & 69.77 & 33.78 \\
      Dense-CLIP \cite{levi_object-centric_2023} & 77.59 & 25.31 \\
      Cluster-CLIP \cite{levi_object-centric_2023} & 78.07 & 29.49 \\
      Ours w/o reranking      & 83.54 & 40.75 \\
      Ours                     & 92.97 & 61.00 \\
      \bottomrule
    \end{tabular}
  }
\end{table}

\begin{table}[!t]
  \caption{Comparison of methods for the RES stage.}
  \renewcommand{\arraystretch}{0.45}
  \label{tab:res}
  \resizebox{0.9\columnwidth}{!}{
    \begin{tabular}{lcc}
      \toprule
      Method & COCO (mAP@50@50) & D$^3$ (mAP@50@50) \\ 
      \midrule
      LISA++ \cite{yang_lisa_2024} & 49.13 & 38.03 \\
      SAM4MLLM \cite{chen_sam4mllm_2025} & 55.48 & 42.16 \\
      APE \cite{shen_aligning_2024} & 71.27 & 41.56 \\
      Ours      & 71.64 & 49.16 \\
      \bottomrule
    \end{tabular}
  }
\end{table}

\begin{table}[!t]
  \centering
  \renewcommand{\arraystretch}{0.45}
  \caption{Comparison of using different region feature extractors for search (w/o reranking).}
  \label{tab:ablation1}
  \resizebox{0.9\columnwidth}{!}{
    \begin{tabular}{lcccccc}
      \toprule
      Method & COCO (mAP@50) & D$^3$ (mAP@50) \\ 
      \midrule
      w/ Cropping+CLIP      & 71.06 & 31.04 \\
      w/ Alpha-CLIP       & 83.54 & 40.75 \\
      \bottomrule
    \end{tabular}
  }
\end{table}

\begin{table}[!t]
  \centering
  \tabcolsep=0.1cm
  \renewcommand{\arraystretch}{0.45}
  \caption{Comparison of different localization approaches.}
  \label{tab:ablation2}
  \resizebox{0.9\columnwidth}{!}{
    \begin{tabular}{lcccccc}
      \toprule
      Method & COCO (mAP@50@50) & D$^3$ (mAP@50@50) \\ 
      \midrule
      Ours (Stage 1)      & 34.19 & 8.84 \\
      Ours       & 71.64 & 49.16 \\
      \bottomrule
    \end{tabular}
  }
\end{table}


\noindent\textbf{Localization approach}: Our stage 1 (Segmentation-aware TIR) can also produce RES results by directly selecting the SAM 2 mask corresponding to the Alpha-CLIP visual embedding with the highest similarity to the query text embedding. Table \ref{tab:ablation2} shows the comparison results. As can be seen, the performance of simply using stage 1 is considerably worse, highlighting the necessity of adopting MLLM and our stage 2 (MLLM-based reranking \& grounding).


\section{CONCLUSION}
In this paper, we introduce MaTIR, a novel task that unifies TIR and RES within a single framework. 
To our knowledge, this is the first study toward this direction. 
We propose a 
pipeline that integrates SAM 2 and Alpha-CLIP for segmentation-aware retrieval, while leveraging an MLLM to refine retrieval results and mask localization. We evaluate our approach on COCO and D$^3$, 
demonstrating its effectiveness in improving both retrieval accuracy and object localization.
The results can serve as baselines for future study.

\bibliographystyle{ACM-Reference-Format}
\balance
\bibliography{refs}

\end{document}